\documentclass[letterpaper, 10 pt, conference]{ieeeconf}  %

\IEEEoverridecommandlockouts                              %

\usepackage{graphics}
\usepackage{epsfig}
\usepackage{mathptmx}
\usepackage{times}
\usepackage{amsmath}
\usepackage{amssymb}
\usepackage{subcaption}
\usepackage{hyperref}

\hypersetup{
    colorlinks=true,
    linkcolor=blue,
    filecolor=magenta,      
    urlcolor=cyan,
}

\renewcommand{\baselinestretch}{0.99}
\usepackage{flushend}

\title{\LARGE \bf
First Steps: Latent-Space Control with Semantic Constraints\\for Quadruped Locomotion
}

\author{Alexander L. Mitchell$^{1,2}$, Martin Engelcke$^{1}$, Oiwi Parker Jones$^{1}$,  David Surovik$^{2}$,\\
Siddhant Gangapurwala$^{2}$, Oliwier Melon$^{2}$, Ioannis Havoutis$^{2}$, and Ingmar Posner$^{1}$
\thanks{$^{1}$Applied AI Lab (A2I), $^{2}$Dynamic Robot Systems (DRS)}%
\thanks{Oxford Robotics Institute (ORI), University of Oxford}%
\thanks{Correspondence to: {\tt\small mitch@robots.ox.ac.uk}}
}

\usepackage{amsmath,amsfonts,bm}

\def\eqref#1{equation~\ref{#1}}

\def\1{\bm{1}}

\def\rvg{{\mathbf{g}}}

\def\rvp{{\mathbf{p}}}
\def\rvq{{\mathbf{q}}}

\def\rvx{{\mathbf{x}}}

\def\rvz{{\mathbf{z}}}

\def\rmS{{\mathbf{S}}}

\def\rmW{{\mathbf{W}}}
\def\rmX{{\mathbf{X}}}
\def\rmY{{\mathbf{Y}}}
\def\rmZ{{\mathbf{Z}}}

\DeclareMathAlphabet{\mathsfit}{\encodingdefault}{\sfdefault}{m}{sl}
\SetMathAlphabet{\mathsfit}{bold}{\encodingdefault}{\sfdefault}{bx}{n}

\begin{document}

\maketitle
\thispagestyle{empty}
\pagestyle{empty}

\begin{abstract}
    Traditional approaches to quadruped control frequently employ simplified, hand-derived models. 
    This significantly reduces the capability of the robot since its effective kinematic range is curtailed.
    In addition, kinodynamic constraints are often non-differentiable and difficult to implement in an optimisation approach.
    In this work, these challenges are addressed by framing quadruped control as optimisation in a structured latent space. A deep generative model captures a statistical representation of feasible joint configurations, whilst complex dynamic and terminal constraints are expressed via high-level, semantic indicators and represented by learned classifiers operating upon the latent space.
    As a consequence, complex constraints are rendered differentiable and evaluated an order of magnitude faster than analytical approaches.
    We validate the feasibility of locomotion trajectories optimised using our approach both in simulation and on a real-world ANYmal quadruped. Our results demonstrate that this approach is capable of generating smooth and realisable trajectories.
    To the best of our knowledge, this is the first time latent space control has been successfully applied to a complex, real robot platform.
\end{abstract}
\section{Introduction}

Four-legged robots (quadrupeds) are capable of traversing a broader range of terrains than wheeled robots and can carry larger payloads with longer battery lives than drones (\cite{ANYmal,robot_terrain_anymal,hyq,cheatah_mini}).
Unlike wheeled robots, legged robots can choose to place their feet on specific locations within the terrain which can support the mass of the robot. This makes them highly suitable for inspection and monitoring tasks in unstructured domains (\cite{robot_terrain_anymal,cheatah_mini,anymal_offshore}). However, this flexibility comes at the cost of platform complexity. 
Kinematics and dynamics are considerably more complicated for quadrupeds than for wheeled robots or drones. This makes trajectory optimisation, i.e. computing feasible paths for robot end-effectors and for the robot's centre of mass (CoM) given kinematic, dynamic, and environmental constraints, one of the central challenges in their deployment (\cite{towr,DarioNlp}).

\begin{figure}[tb]
    \centering
    \includegraphics[width=1.0\linewidth]{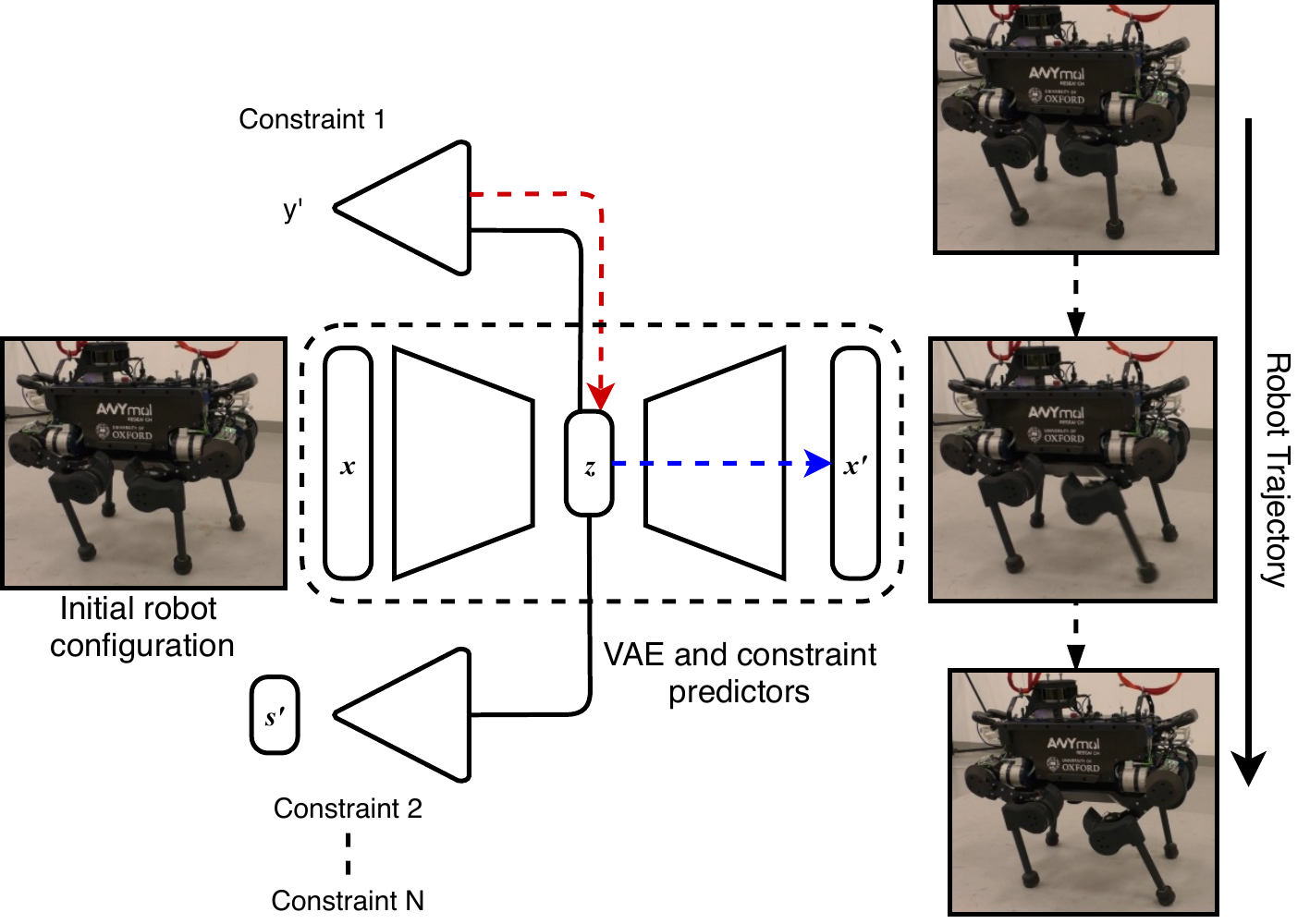}
    \caption{A VAE encodes the robot state and captures correlations therein in a structured latent space. Once trained, performance predictors (triangles) attached to the latent space predict if constraints are satisfied and apply arbitrarily complex constraints such as robot stability. The flexibility of this approach allows for the application of any number of constraints by performing activation maximisation of a loss summed over all active constraints (see, for example, Eq.\ \ref{eq:loco_loss}). Decoding positions in latent space (blue) along the optimisation trajectory translates to movement of the robot.}
    \vspace{-0.5cm}
    \label{fig:anymal_step}
\end{figure}

Traditionally, trajectory optimisation for quadrupeds is solved using constrained optimisation. 
However, the robot’s feasible joint space and dynamics such as stability, torque limits, and contact forces, require complex often non-differentiable constraints. This makes the optimisation intractable.
A typical approach, therefore, is to use approximate, hand-derived dynamic models and arbitrarily reduce the kinematic range of the robot \cite{towr}.
These linear approximations typically over-simplify the problem and limit platform capabilities, leading to overly narrow convergence basins given feasible initial states (e.g. \cite{towr,towr++,fastTraj}).
As a result, trajectories solved using linear approximations are only suitable for specific use-cases and do not easily generalise to novel situations.
In contrast, approaches which account for the full kinodynamic representation of the robot require iterative, sequential optimisations to check for constraint satisfaction \cite{static_stability}. This often makes them computationally infeasible for real-time deployment.

\begin{figure*}[t]
  \centering
  \includegraphics[width=.75\textwidth]{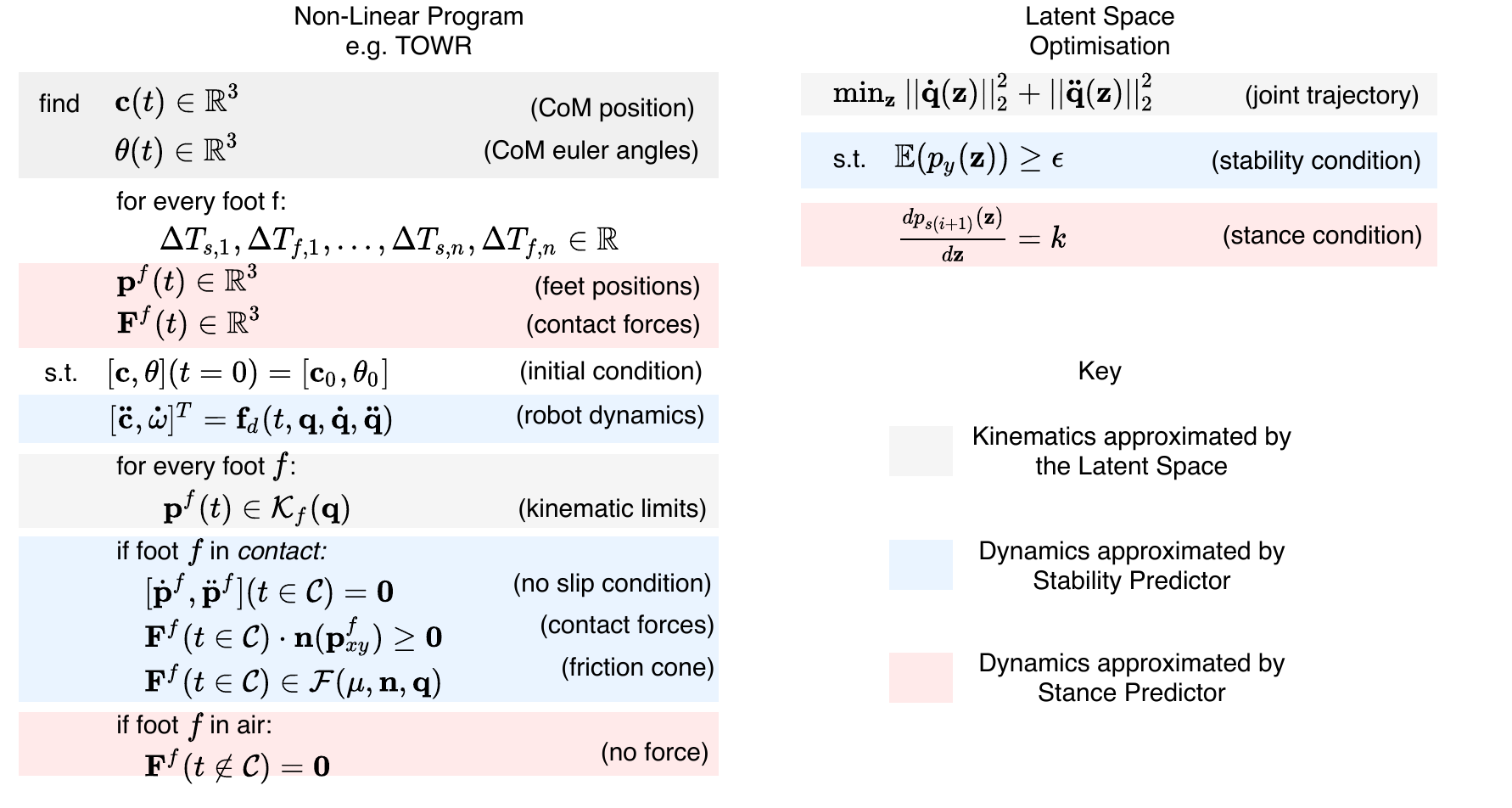}
  \caption{Conceptual comparison of solving a non-linear program (NLP) such as TOWR with latent space optimisation. The NLP splits locomotion into an optimisation subject to a series of constraints, whilst trajectories in our approach are solved via gradient descent in a structured latent space. Complex constraints such as stability are enforced using learned classifiers (performance predictors) and gradients from differentiating a target loss are used to update a trajectory in latent space. Performance predictors replace multiple and often non-differentiable dynamics exhibited in the NLP.}
  \label{fig:optimisation_comparison}
  \vspace{-0.5cm}
\end{figure*}

Here, we propose a radically different approach to quadruped control. Using a generative model of the robot state we perform trajectory optimisation by directly optimising the position in a structured latent space, which captures a statistical model of the robot’s feasible joint-space (see, Fig.~\ref{fig:anymal_step}). In particular, we employ a \emph{variational autoencoder (VAE)} (\cite{vae,vae_1}) to encode the robot state including  joint positions. Inspired by \cite{imagine_that}, operational constraints are induced via \emph{high-level}, semantic indicators and represented by learned performance predictors operating directly in the latent space. Arbitrarily complex constraints -- and goals -- can thus be enforced by performing gradient-based optimisation in the latent space driven by the performance predictors via activation maximisation \cite{activation_maximisation}.

By implicitly capturing a full kinodynamic model, our approach enables \emph{efficient} trajectory optimisation without relying on hand-specified dynamics models or linear approximations. In contrast to prior art, it predicts whether constraints are satisfied with a single pass through our neural network. More importantly, it enables direct optimisation of traditionally non-differentiable dynamics (see, Fig.\ \ref{fig:optimisation_comparison}) 
together with the correlations captured by the latent space, and this results in inherently smooth, feasible robot trajectories when decoding poses into state-space along the optimised path in latent space.

We validate our approach to latent-space control in the context of quadruped locomotion and constrain robot stability and end-effector contact dynamics to obtain a walking gait (Fig.\ \ref{fig:anymal_step}). Using both simulated and real robot experiments, we demonstrate that our approach to latent-space control is able to find trajectories from unstable to stable robot configurations as well as to execute realisable gait cycles. This is achieved with execution times almost an order of magnitude faster than a comparable analytical approach \cite{feasible_region}.

To the best of our knowledge this is the first approach to latent-space control for a complex dynamic system which is validated both in simulation and on a real platform.

\section{Related Work}

As torque controlled robots have become more complex, the weaknesses of traditional trajectory optimisation approaches have become more pronounced. These mostly stem from large configuration-spaces and the complexity of model dynamics.
Hand-derived dynamics models such as centroidal dynamics \cite{centroidal_dynamics} can be used as dynamic constraints (\cite{DarioNlp,towr}).
For example, \emph{TOWR} \cite{towr} is a general approach for solving locomotion tasks for any legged robot over known terrain.
Specifically, TOWR solves a non-linear program for the centre of mass and feet trajectories whilst optimising the contact forces.
This is computationally slow and prohibits online updates to the current plan to react to environmental changes.
Furthermore, kinematic constraints restrict the range of the robot's movement.
While TOWR is suitable for highly-dynamic locomotion, our approach is conceived to be sufficiently general for low-velocity operation such as locomotion over rough terrain.

\cite{feasible_region} estimates robot stability by finding a ``margin of stability'' while taking into account the robot's centre of mass, friction forces, and torque limits.
We use this formulation to create training data for our stability predictor and as a performance baseline. 
While the original approach requires sampling and evaluating multiple points, our method directly optimises for stable trajectories using a learned stability predictor.

Reinforcement learning (RL) (e.g. \cite{Jemin,gangapurwala2020guided}) is a promising alternative to optimisation-based approaches.
However, RL policies are difficult to train, requiring vast quantities of data, and suffer from unstable gradient estimates.
Once trained, specific constraints such as foothold placements are impossible to enforce.
In contrast, our approach is flexible: additional constraints are applied by constructing an appropriate loss function summed over all active constraints, which is differentiated to provide a gradient update.

Recent advances in latent-space optimisation also attempt to overcome the limitations of traditional approaches.
Universal Planning Networks (UPN) \cite{UPN} use a gradient-based method in a structured latent space to find trajectories between an initial and final image. 
Similarly, Embed to Control (E2C) \cite{embed2control} performs system identification and state estimation from visual inputs for control problems.
In addition, a forward dynamics model is learned to traverse the latent space given an input.
Both UPN and E2C need to be provided with a goal condition.
In contrast, performance predictors, presented here, are capable of predicting if arbitrarily complex constraints, such as robot stability, are satisfied. 
A target loss is evaluated using these performance predictors and differentiated until the latent variable is guided into the terminal set via gradient descent.

\section{Gradient-Based Control in Latent Space with Performance Predictors}

\subsection{Learning a Latent Representation of Robot States}\label{section:training_vae}

VAEs learn compact and smooth representations of high-dimensional data (\cite{vae,vae_1}).
We utilise a VAE to encode the robot's joint angles $\rvq$, feet positions in the base frame $\rvp^f$ (where $f\in\mathbb{N}$ is the number of feet), joint torque $\mathbf{\tau}$, contact forces $\mathbf{\lambda}$, and the gravity body force vector $\rvg$, forming the input to the VAE ${\rvx=[\rvq, \rvp^f, \mathbf{\tau}, \mathbf{\lambda}, \rvg]^T}$, ${\rvx \in \mathbb{R}^{51}}$.
To find these quantities, \emph{static}, snapshots of the robot in random, but feasible joint configurations are sampled under the condition that joint velocities and accelerations are set to zero.
Finally, the VAE is trained to reconstruct the input $\rvx$ via a latent space denoted $\rvz \in \mathbb{R}^{N_z}$.
Training is achieved by minimising the KL-regularised mean-squared error of the reconstruction:
\begin{align}\label{eq:loss_vae}
    \mathcal{L}_{VAE} = MSE(\rvx, \rvx') + \beta D_{KL}[q(\rvz|\rvx) || p(\rvz)]
\end{align}

Given the importance of robot stability in general and information about its current stance for the purpose of locomotion in particular, the VAE is trained jointly with two \emph{performance predictors} \cite{imagine_that} which estimate the probability of the robot being stable $y'$ and the probability that it is currently in a specified stance $s'_i$ using a binary cross-entropy (BCE) loss.
Gradients from these two classifiers are backpropagated through the VAE encoder and aid in shaping the latent representation.
Given two hyperparameters $\mu_1$ and $\mu_2$ for weighting the loss terms, the full training objective can thus be written as:
\begin{align}\label{eq:loss_full}
    \mathcal{L} = \mathcal{L}_{VAE} + \mu_1 BCE(y, y') + \mu_2 BCE(s_i, s'_i)
\end{align}\label{eq:losses_1}

\begin{figure}[h]
  \centering
  \includegraphics[width=0.45\textwidth]{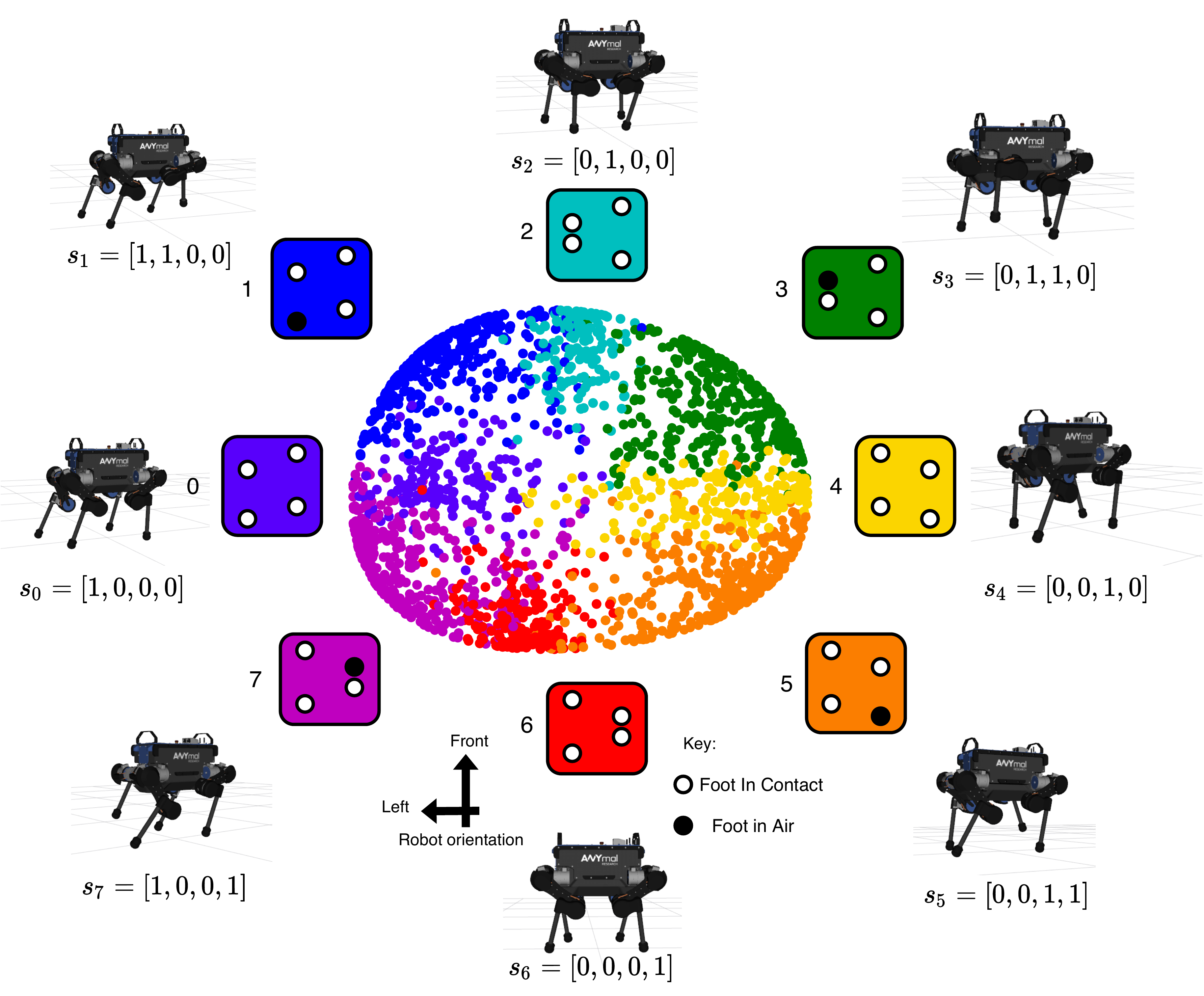}
  \caption{The latent space (central panel) is structured and clearly shows the stance clusters (coloured).
  For visualisation, principle component analysis (PCA) \cite{pca} with a radial basis function (RBF) kernel reduces the latent space from 64 dimensions to two.
  Walking is achieved by cycling clockwise through the eight stances, which are represented by coloured diagrams orientated so that forward is up the page and left aligns with reader's left. 
  Open white circles represent the target position of feet in contact with the ground; closed black circles represent feet in the air.}
  \label{fig:latent_space_structuring}
  \vspace{-0.5cm}
\end{figure}

This provides an inductive bias to the model, which structures our latent space so that consecutive stance clusters are adjacent in the latent space (see, Fig.~\ref{fig:latent_space_structuring}). 
This structure is achieved by the design of the stance labels.
Stances in the walk gait where four feet are on the ground are one-hot encoded, whilst intermediary classes with a foot is in the air are two-hot encoded.
Therefore, regions exist in latent space where two classifiers are active and, subsequently, consecutive stance clusters are encoded in sequence in the latent space.

\subsection{Constrained Latent-Space Traversal}\label{section:am}

The performance predictors estimate whether non-linear and often non-differentiable kinodynamic constraints, as well as goal conditions are satisfied. Importantly, formulating these constraints as neural network classifiers renders them differentiable with respect to the latent variable $\rvz$.
This allows us to update a latent variable sequentially via gradient descent until a constraint is satisfied.
This is performed either to steer a trajectory to satisfy a terminal constraint or to enforce a condition upon the entire trajectory.
Using gradient-descent to update a latent variable while a performance predictor enforces a particular constraint is a use of \emph{activation maximisation} (AM) \cite{activation_maximisation}.

The general case for \emph{activation maximisation} requires one or more differentiable loss functions $\mathcal{L}_k(y, y')$, where $y$ is the target value and $y'$ is the predicted value.
Each loss function has an associated step size $\alpha_k$, which scales the respective gradient step.
Finally, an update to the latent variable $\rvz$ is made via gradient descent in the latent space:
\begin{equation}\label{eq:am}
    \rvz \leftarrow \rvz - \nabla \sum_k \alpha_k \mathcal{L}_k(y, y')  
\end{equation}

This approach is inherently flexible as additional constraints are applied simply by adding additional loss functions prior to a gradient update.

\subsection{Activation maximisation for stability}\label{section:am_stability_methods}

A specific use of constrained latent-space traversal in the context of quadruped control is disturbance rejection.
Performance predictors are not only capable of predicting if the robot has become unstable, but also finding smooth and stabilising trajectories by continuously backpropagating the error from a performance predictor back into the model's latent representation $\rvz$ using activation maximisation (AM).
Each latent update $\rvz$ is decoded to state space $\rvx'$ to create a trajectory.
By choosing to minimise the BCE loss, the predicted probability that the robot is stable $y'$ is driven close to the desired probability $y$.
Once a step size $\alpha_y$ is selected, AM for robot stability is defined:
\begin{equation}\label{eq:am_stable_loss}
    \rvz \leftarrow \rvz - \nabla \alpha_y BCE(y, y')
\end{equation}

The resulting trajectories are evaluated for their efficacy and feasibility in section \ref{section:experiments}.

\subsection{Optimisation for Locomotion}\label{section:optimisation_methods}

Locomotion requires finding a trajectory which allows the quadruped to take a series of steps through a gait sequence while the robot remains stable.
The robot is stable when the feet in contact support the CoM without slipping while respecting maximum torque limits.
To solve for a walk trajectory in latent space, two performance predictors are required: one for stability and another to move the robot between gait stances (see Fig.~\ref{fig:latent_space_structuring}).

To frame locomotion as an optimisation problem, we minimise the joint velocities $\dot{\rvq}$ and accelerations $\ddot{\rvq}$ over an entire trajectory. The joint velocities and accelerations are calculated from the joint angles using a forward-Euler derivative estimation. Intuitively, minimising these values will result in smoother, more readily executable trajectories as the actuators perform best when the input is varied smoothly \cite{SEA}. Therefore, we solve for a trajectory that minimises the squared $L_2$-norm of the joint velocity and acceleration of the robot, whilst maximising the robot's stability $y$ and linearly increasing the probability of being in the next stance $s_{i+1}$. 
Trajectories are of length $N$, which is $N/f_s$ seconds long, where $f_s$ is the sampling frequency. 
The initial robot configuration $\rvx_0$ is encoded into the latent space to find $\rvz_0$. 
$N$ repeats of $\rvz_0$ are stacked to create $\rmZ \in \mathbb{R}^{N,N_z}$, and this forms the latent-space trajectory.
Concretely, we perform
\begin{align}\label{eq:locomotion_optimisation}
    \rmZ^* = \underset{\rmZ}{\arg \min} \quad \bigg( &|| \dot{\rvq}(\rmZ) ||_2^2 + || \ddot{\rvq}(\rmZ) ||_2^2 \bigg) \\
    s.t. \qquad & \mathbb{E}(p_y(\rmZ)) \geq \epsilon \label{eq:stable_constra}\\
               & \frac{d p_{s(i+1)}(\rmZ)}{d \rmZ} = k \label{eq:stance_constra}
\end{align}

The constraints in Equations \ref{eq:stable_constra} and \ref{eq:stance_constra} are converted to costs using Lagrange multipliers $\lambda_i$ and added to the objective function in Equation \ref{eq:locomotion_optimisation} to form the locomotion loss 
\begin{equation}\label{eq:loco_loss}
    \mathcal{L}_{loco} = ||\dot{\rvq}||_2^2 + ||\ddot{\rvq}||_2^2 + \lambda_0 BCE(y, y') + \lambda_1 BCE(s, s')
\end{equation}

$\rmZ$ is passed through the VAE's decoder to obtain the reconstructed $\rmX$, and also through the performance predictors to provide stability ($\rmY$) and stance ($\rmS$) estimates.
This output is concatenated into $\rmW = [\rmX, \rmY, \rmS]$ and multiplied with a selector matrix to compute the quantities needed to evaluate the locomotion loss $\mathcal{L}_{loco}$.
$\mathcal{L}_{loco}$ is summed over the entire trajectory and multiplied with a step size $\alpha_{loco}$ before being differentiated with respect to $\rmZ$
to provide a gradient update such that

\begin{align}\label{eq:loco_update}
    \rmZ \leftarrow \rmZ - \nabla_{\rmZ} \alpha_{loco} \mathcal{L}_{loco}(\rmZ)
\end{align}

\section{Experiments}\label{section:experiments}
We apply constrained latent-space traversal to solve complex planning tasks on a quadruped robot. In particular, we focus on stability and locomotion. Our evaluation first investigates whether our approach to latent-space control is able to provide feasible trajectories when rapid robot stabilisation is required. This is compared to a baseline of performing activation maximisation by backpropagating gradients directly into the input of a feed-forward classifier rather than a structured latent encoding. Secondly, we add a constraint on the robot stance to enable it to take steps while remaining stable throughout the movement. 
We first describe the dataset of sampled robot configurations and associated ground-truth labels required for training the VAE and the performance predictors. We then provide the architectural details of our networks before presenting results of the individual experiments.

\subsection{Dataset Generation}

A dataset is sampled consisting of triplets $\{\rvx, y, s\}$, where $\rvx$ is a static robot configuration and $\{y$, $s\}$ are the associated ground truth labels for robot stability (binary) and stance (one out of eight), respectively.
The robot's centre of mass location is sampled uniformly within $\pm120\mathrm{mm}$ in the longitudinal direction and $\pm100\mathrm{mm}$ in the lateral direction.
Feet not in contact with the ground are sampled at heights of $\{0,40,80,120\}\mathrm{mm}$.
We collect a total of 88,000 triplets whereby 70,400 are used for training (80\%) and 17,600 for testing (20\%).

\textbf{Stability:}
To obtain the ground truth labels for robot stability, we follow the approach in \cite{feasible_region}, which considers friction limits of the ground and the robot as well as torque limits of the series elastic actuators \cite{SEA} of the robot.
The friction cone model estimates if the robot's feet in contact will slip on the surface.
A foot in contact is defined such that the velocity or acceleration of the foot is zero.
This means that there are different sets of contact forces and torques for a foot in contact with a surface and a foot which touches the surface without being in contact.
The coefficient of friction is assumed to be constant and hence as long as the lateral forces are less than the force normal to the surface multiplied by the coefficient of friction, the robot will not slip.

\textbf{Stances:}
Robot stances are defined by which feet are in contact with the ground and the relative ordering of the feet.
Specifically, we break up a walking gait into eight phases where each phase is associated with a unique stance (see Fig.\ \ref{fig:latent_space_structuring}).
The eight stances are made up of four stances with one foot in the air and four with all four feet in contact.
Locomotion thus corresponds to cycling through these stances, see (Fig.\ \ref{fig:latent_space_structuring}).

\subsection{Architecture Details}\label{section:architecture}

The VAE architecture comprises an encoder and decoder, which each have two fully-connected layers with 256 units and ELU non-linearities \cite{elu}.
The latent space is of width 64 with a zero mean and identity variance Gaussian prior, and a diagonal Gaussian posterior distribution. The stability classifier consists of three fully-connected layers with a width of 64 units and ELU non-linearities. 
It has a single output for the stability semantic label. %
The stance classifier consists of three fully-connected layers with a width of 64 units and ELU non-linearities. It predicts a four element output vector providing the stance encoding (see Section \ref{section:training_vae}). 
In the case of the locomotion experiments described below, our model takes the position of the robot's centre of gravity as an additional input. This extra input is reconstructed along the trajectory and was found useful for debugging purposes. We have verified empirically that it does not significantly impact model performance otherwise.
As a baseline, a second feed-forward neural network is trained so that activation maximisation is performed directly into input space as opposed to the latent space.
This classifier network takes as input the robot data $\rvx$ and predicts the probability that the robot is stable and which stance the robot is in.
The architecture used consists of three fully-connected layers with a width of 64 units and ELU non-linearities.

\subsection{Stability Experiments}

The context for this experiment is disturbance rejection. 
If, for example, the robot is perturbed and no longer stable, we demonstrate how activation maximisation (AM) in a suitable latent space can find a smooth trajectory to a stable pose.
At each time step, we backpropagate the stability error from a performance predictor back into the model's latent representation $\rvz$. This is decoded to $\rvx'$ and sent as a command to the robot's low-level controller until the robot is stable.  
We compare AM into a structured latent space (\emph{latent-AM}) with applying AM directly to the input (\emph{input-AM}). 
The latter scheme bypasses the latent space and, therefore, input-AM does not explicitly account for any correlations between the input state variables.

In each experiment (latent-AM and input-AM), we randomly initialise 1000 unstable configurations. 
AM is then driven by a binary cross entropy loss with step size ${\alpha_y=1\times10^{-3}}$; and $\rvz$ is updated following Eq.~\ref{eq:am_stable_loss} for 300 gradient steps. 

Our measure of success is the proportion of episodes that feasibly transition from an unstable to a stable configuration. 
Here we define a stable configuration in terms of the robot's CoM together with the contact points where its feet meet the ground. 
Hence, it is necessary for both models to also predict which feet are in contact with the ground. 
Given that these contact points define the vertices of a polygon projected on the ground, the robot is stable when its CoM hovers within the support polygon. 
Otherwise, if the CoM projection is outside of this polygon, the robot is overextended and considered unstable. 
Since the robot's CoM can be inferred from its joint configurations and feet positions in the base coordinates, the CoM was not included as an input to the model.

Furthermore, we only consider an experiment a success if the robot configurations which constitute a trajectory are \emph{kinematically} feasible.
This means that there are no self-collisions or joint limit violations in all the poses which constitute a trajectory.

We report that the overall proportion of latent-AM experiments that successfully reach a stable pose is 94.6\%.
However, input-AM success is significantly lower: only 62.6\% of input-AM experiments succeeded.

The performance of both latent-AM and input-AM is compared once 95\% confidence intervals are inferred using the Wilson score interval \cite{wilson}.
Latent-AM confidence intervals range from 93.1\% to 96.2\%, whilst input-AM varies from 59.5\% to 65.5\%.
There is no overlap between latent-AM and input-AM confidence bounds, and, hence, there is a significant and substantial difference in performance between the two schemes.

To probe the robustness of the final robot pose, we consider more restrictive definitions of stability which shrink the support polygon, to infer a \emph{stability margin}.
Conceptually, CoMs near the centre of the support polygon are more stable than CoMs near the perimeter. 
The stability margin is defined as the area with which the support polygon can be reduced while the robot remains stable. 
In practice, trajectory optimisers shrink the support polygon by $\sim$20\% to mitigate any error in estimating the CoM position due to sensor and actuator noise \cite{robust_TO}.
Table \ref{table:success_results} shows the proportion of experiments which find a stable final pose as the support polygon shrinks in increments $\sim$6\% from its original periphery (no change in the margin) up to 31\% smaller polygons.
The latent-AM success-rates never drop below 92\%, while input-AM performance deteriorates considerably from 63\% to 52\%.
This clearly indicates that latent-AM is significantly more likely to find a robustly stable robot pose.

\begin{table}[h]
\vspace{-0.1cm}
\centering
\caption{The percentage of experiments which terminate at a stable robot configuration for both latent-AM and input-AM. 95\% confidence intervals are also presented in brackets and calculated using Wilson score interval \cite{wilson}. To measure the robustness of this stable pose, the support polygon is shrunk by a margin from zero (no margin) to 31\% in 6\% intervals. With no margin, 95\% of latent-AM trajectories reach a stable pose as opposed to only 63\% of input-AM plans. The success of both schemes reduces as the margin increases, but input-AM performance degrades severely ($\sim10$\%). However, latent-AM success remains above 92\%.
}
\label{table:success_results}
\begin{tabular}{c c c}
\multicolumn{1}{c}{}  & \multicolumn{2}{c}{Successful Trajectories/ \%} \\ %
\multicolumn{1}{c}{ Margin/ \%}  &\multicolumn{1}{c}{\bf Latent AM}  &\multicolumn{1}{c}{\bf Input AM}
\\ \hline \hline
0  & \textbf{94.6} (91.9, 95.3) &  62.6 (59.6, 65.5) \\
7  & \textbf{93.8} (91.2, 94.8) &  59.8 (56.7, 62.8) \\
13 & \textbf{93.2} (90.9, 94.5) &  58.4 (55.3, 61.4) \\
19 & \textbf{92.9} (90.6, 94.3) &  56.4 (53.3, 59.4) \\
25 & \textbf{92.7} (90.6, 94.3) &  54.8 (51.7, 57.9) \\
31 & \textbf{92.6} (90.5, 94.2) &  52.0 (48.9, 55.1) \\
\end{tabular}
\end{table}

\begin{figure*}[h]
    \centering
    \includegraphics[width=0.99\textwidth]{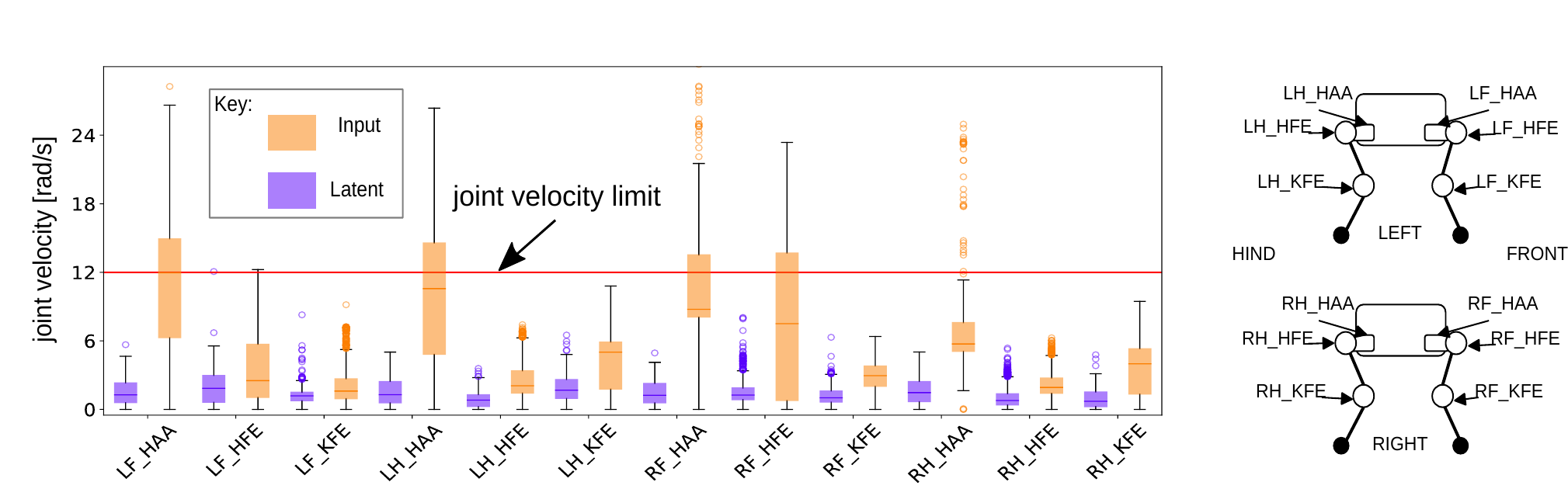}
    \caption{A joint trajectory is only feasible if maximum permissible joint velocity (12 rad/s) is not exceeded. The proportion of input-AM trajectories which are infeasible is 59.3\%, which contrasts latent-AM where no experiments violate the maximum joint velocity. The latent-AM median-maximum velocity is always lower than that of the input-AM results.}
    \vspace{-0.5cm}
    \label{fig:joint_vels}
\end{figure*}

For a trajectory to be realisable, we emphasise that the robot trajectory must also be \emph{dynamically} feasible.
By dynamically feasible we mean that the simulation cannot move in ways that would be physically impossible for a real robot. 
Such simulated trajectories may end up at a stable configuration, but, because they cannot be implemented, are uninteresting from the perspective of a real robot. 
We consider an episode infeasible if the simulated trajectory exceeds the maximum joint velocity of 12 rad/s, as reported in the ANYmal motor specifications \cite{ANYmal}.
For replicability, we note that our trajectories are sampled at the target control frequency of 200 Hz.

Fig. \ref{fig:joint_vels} displays the maximum joint velocities from all the experiments, and input-AM joint velocity violations are visible.
Furthermore, no latent-AM experiments exceed the joint velocity limit and therefore all successful trajectories are feasible.
In contrast, only  59.3\% of all input-AM trajectories were feasible with 95\% confidence interval of (56.3\%, 62.3\%).
Crucially, once success and feasibility are considered together, only 25.9\% of input trajectories, with a 95\% confidence interval of (22.9\%, 28.4\%) are practically useful for use on the robot.
This compares to 94.6\% feasibly successful latent trajectories.

\begin{figure}[h]
    \centering
    \includegraphics[width=0.49\textwidth]{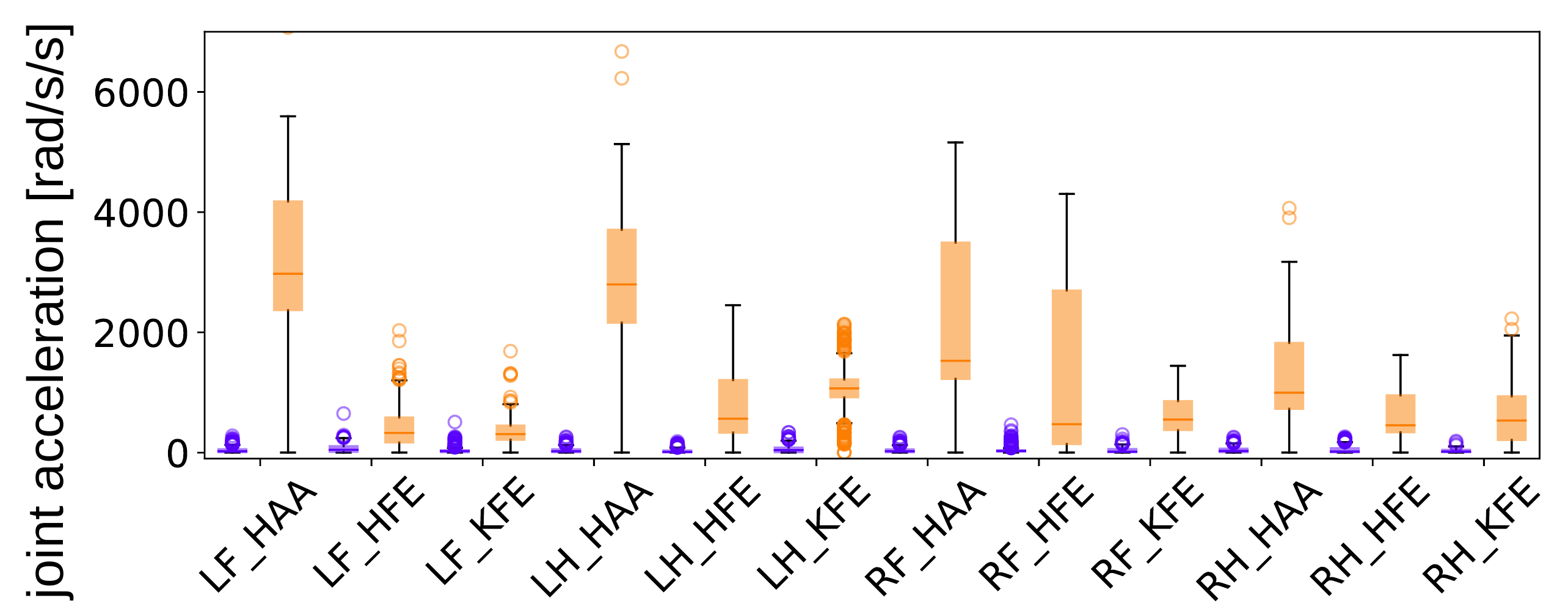}
    \caption{The maximum joint acceleration serves as a metric of smoothness. Jerky trajectories will unsettle the robot and can cause it to fall over. Here we see that latent-AM trajectories are far smoother whilst exhibiting a narrower spread of absolute accelerations than the input-AM ones.}
    \vspace{-0.2cm}
    \label{fig:joint_accels}
\end{figure}

Smooth trajectories are essential to avoid unsettling, tipping over, or even damaging the robot. 
We find that the latent-AM trajectories are significantly smoother than input-AM trajectories, as shown by the derivatives of each joint velocity in Fig.\ \ref{fig:joint_accels}.

Analysis of the latent space show that four latent dimensions have a lower variance than the prior distribution.
In contrast, the other latent units have a mean close to zero and a near unit variance, which closely map the prior distribution.
Interestingly, only the four low-variance latent dimensions are updated by latent-AM, which further indicates that the latent space is exploiting the correlations which exist in the input, and that the effective latent space is four units.

\begin{figure}[h]
    \centering
    \vspace{-0.3cm}
    \includegraphics[width=0.45\textwidth]{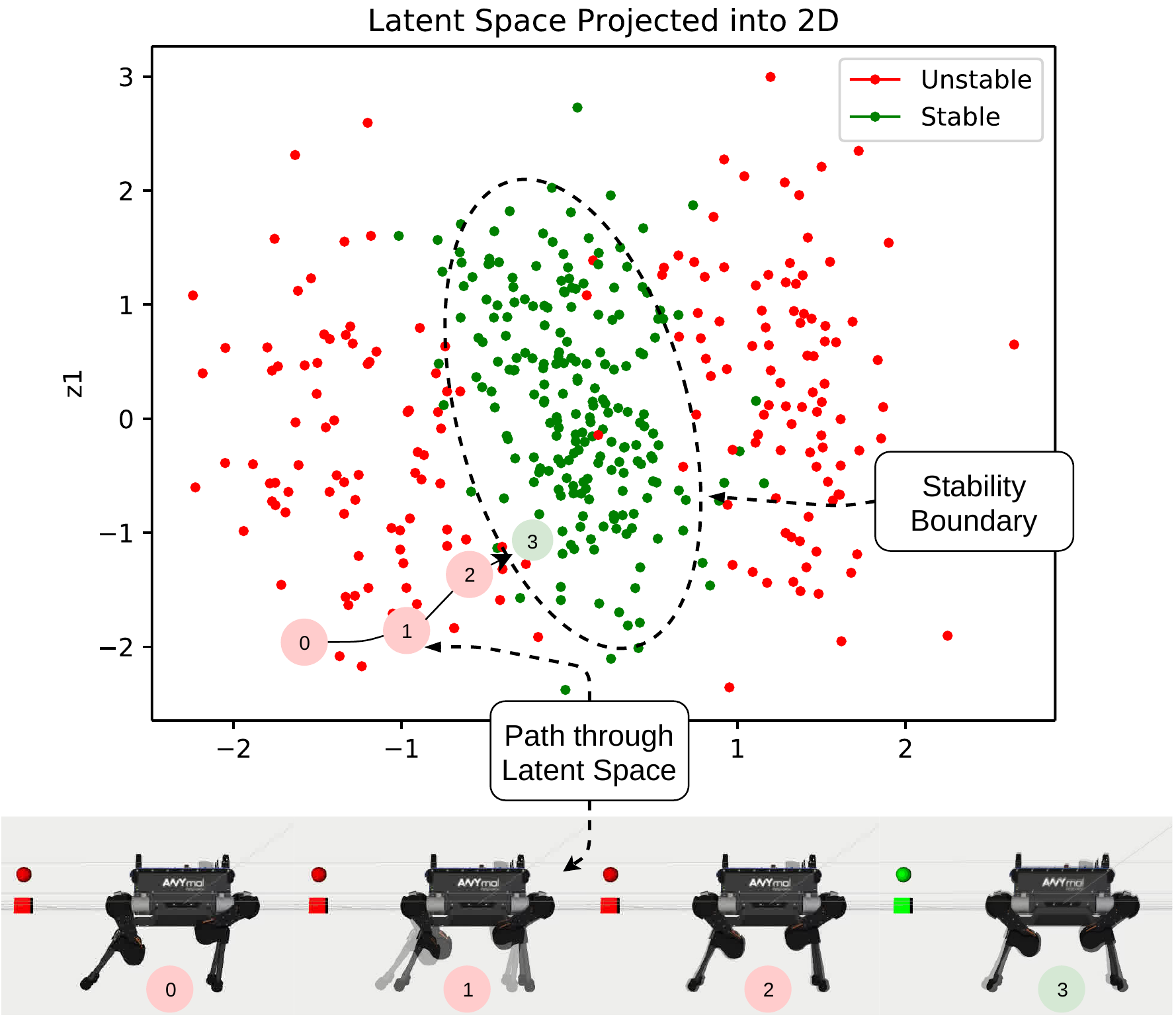}
    \caption{We project the latent space down to 2D via PCA with an RBF kernel. Here we can see clustering of Stable robot configurations in green surrounded by unstable configurations in red. The robot configuration in pose 0 is initially unstable. Activation maximisation provides a smooth and physically realisable trajectory from this unstable pose to a stable one.}
    \label{fig:latent_space_2d}
    \vspace{-0.2cm}
\end{figure}

Further investigation shows that the latent space is divided into stable and unstable clusters.
Fig.~\ref{fig:latent_space_2d} displays stable robot poses (green) surrounded by unstable configurations (red), along with a representative latent trajectory which moves from an unstable robot pose to a stable one.
Below this are snapshots from a decoded trajectory sampled from a latent-AM trajectory, showing a smooth movement of the CoM to a stable position.
In conclusion, the clustering of stable poses coupled with the correlations captured by the latent space could be the causes for the smoothness exhibited by the latent-AM trajectories.

\begin{figure*}[tp]
\begin{subfigure}[t]{.16\linewidth}
\centering
\includegraphics[width=\textwidth]{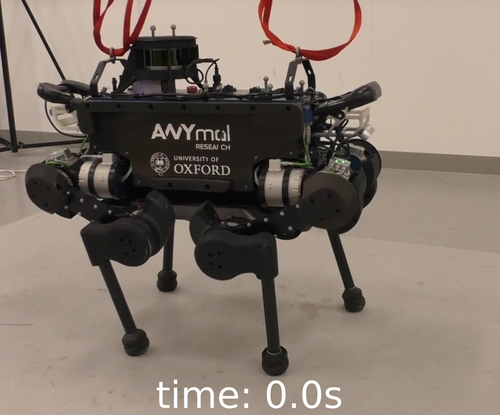}
\end{subfigure}
\hfill
\begin{subfigure}[t]{.16\textwidth}
\centering
\includegraphics[width=\textwidth]{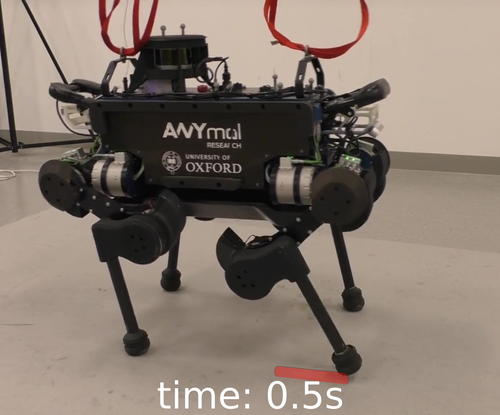}
\end{subfigure}
\hfill
\begin{subfigure}[t]{.16\linewidth}
\centering
\includegraphics[width=\textwidth]{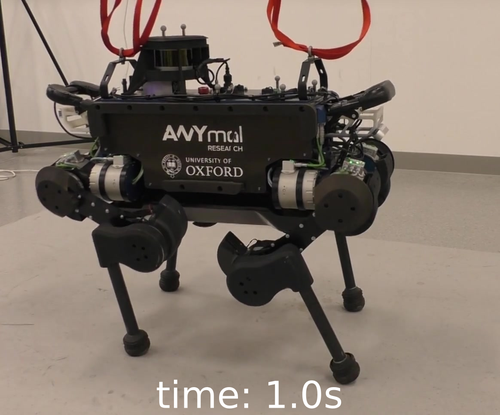}
\end{subfigure}
\hfill
\begin{subfigure}[t]{.16\textwidth}
\centering
\includegraphics[width=\textwidth]{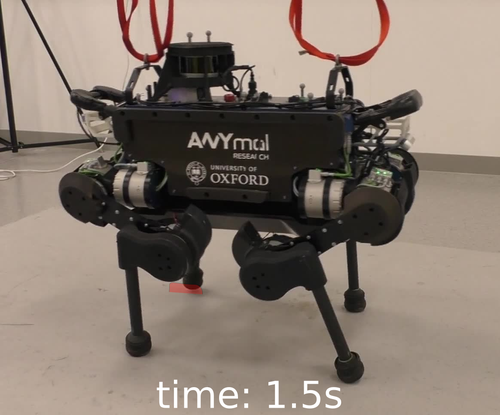}
\end{subfigure}
\hfill
\begin{subfigure}[t]{.16\linewidth}
\centering
\includegraphics[width=\textwidth]{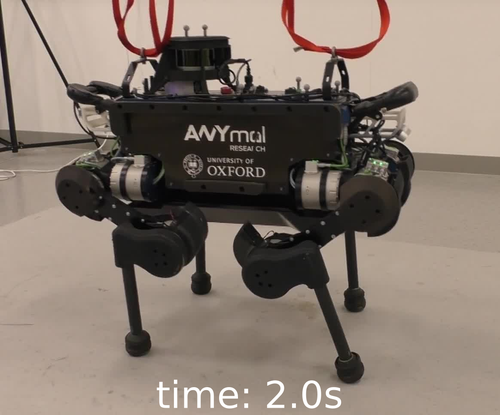}
\end{subfigure}
\hfill
\begin{subfigure}[t]{.16\textwidth}
\centering
\includegraphics[width=\textwidth]{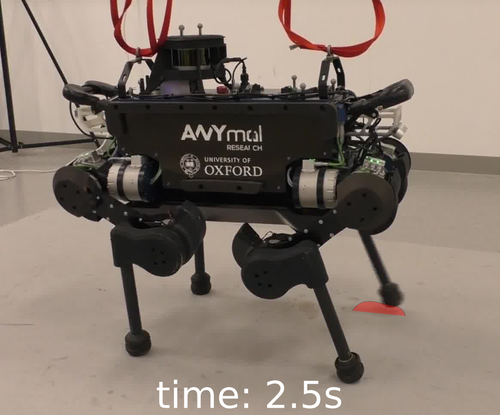}
\end{subfigure}
\newline
\begin{subfigure}[t]{.16\linewidth}
\centering
\includegraphics[width=\textwidth]{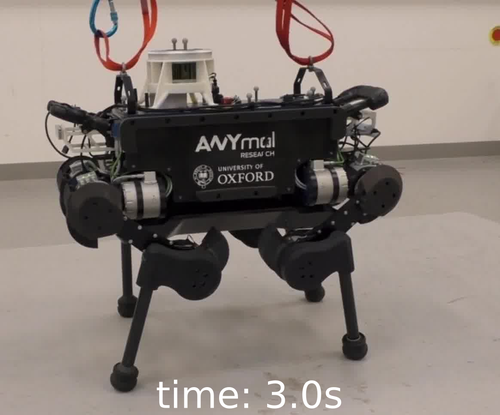}
\end{subfigure}
\hfill
\begin{subfigure}[t]{.16\linewidth}
\centering
\includegraphics[width=\textwidth]{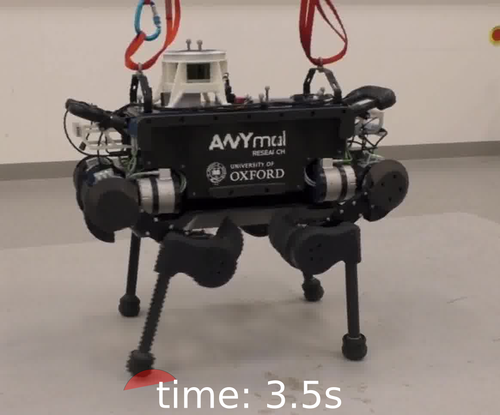}
\end{subfigure}
\hfill
\begin{subfigure}[t]{.16\linewidth}
\centering
\includegraphics[width=\textwidth]{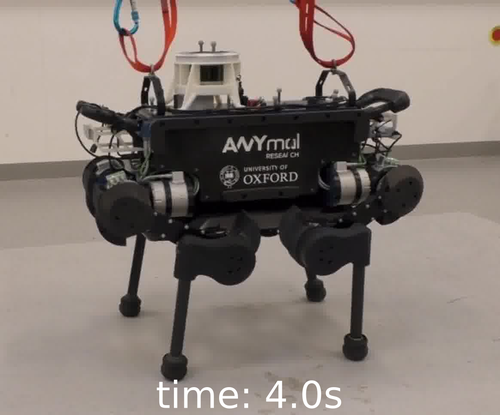}
\end{subfigure}
\hfill
\begin{subfigure}[t]{.16\linewidth}
\centering
\includegraphics[width=\textwidth]{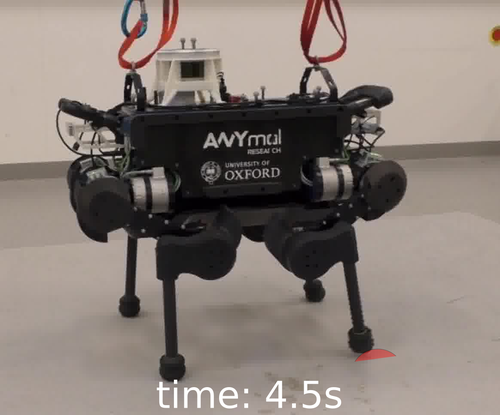}
\end{subfigure}
\hfill
\begin{subfigure}[t]{.16\linewidth}
\centering
\includegraphics[width=\textwidth]{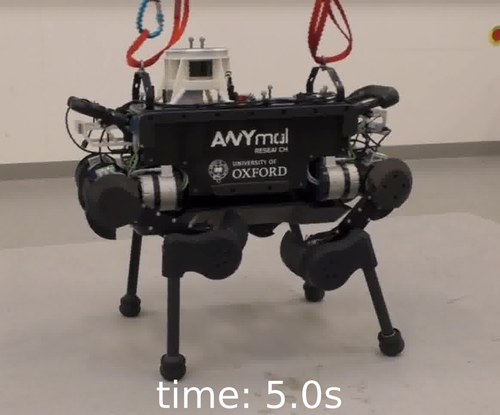}
\end{subfigure}
\hfill
\begin{subfigure}[t]{.16\linewidth}
\centering
\includegraphics[width=\textwidth]{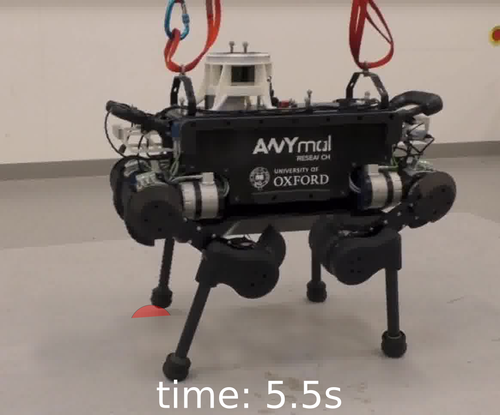}
\end{subfigure}
\caption{The first six steps of the locomotion trajectory solved via optimisation in the latent space on the real robot. Here two separate runs with differing camera angles are displayed together. The swept area of each foot swing is displayed in red. 
A full video of the robot walking is found \href{https://youtu.be/z1UopkfIPhI}{here}.}
\vspace{-0.5cm}
\label{fig:anymal_walking}
\end{figure*}

Finally, we note that evaluating a stability criterion takes 0.72ms to compute whilst a comparable analytical solution in \cite{feasible_region} takes 6.56ms.
This comparison uses an Nvidia Quadro P2000 on a laptop with an Intel Xeon(R) 2.9GHz CPU.
This improvement in performance is attributed to the analytical solution requiring a series of sequential optimisation, as opposed to a single pass through a neural network on a GPU.
Our approach, therefore, is more suitable for applications which require closed loop, online planning.

\subsection{Locomotion}\label{section:locomotion_exps}

Having shown how AM is used to drive a simulated quadruped to find a smooth path that rejects unwanted perturbations, we show that an extended framework, which optimises over an entire trajectory is used to control other kinds of movement. 
In particular, we demonstrate that the addition of performance predictors for idealised stance configurations can be used to make a real ANYmal robot walk~\footnote[3]{A video of the locomotion trajectory executed on the robot is found at \url{https://youtu.be/z1UopkfIPhI}.}.

In our approach, the stability and stance performance predictors are sufficient to ensure that the locomotion trajectory is feasible and realisable.
The overall locomotion cost is described by Eq.~\ref{eq:loco_loss}. 
It consists of a quadratic cost on the joint velocities and accelerations and of BCE losses. 
The BCE losses constantly encourage the stability of the trajectory while linearly increasing the probability of being in the next stance.
For the walking demonstration, a horizon of $N=400$ is chosen together with a step size $\alpha_{loco}$ of $2\times10^{-4}$ in Eq.~\ref{eq:loco_update}, while the Lagrange multipliers in the locomotion loss (Eq.~\ref{eq:loco_loss}) are both set to ${\lambda_0=\lambda_1=10^2}$.

Snap-shots of a complete walk cycle executed on the real ANYmal robot are presented in Fig.\ \ref{fig:anymal_walking}.
The controller in \cite{DarioNlp} tracked our locomotion trajectory with a tracking error of ($1.82\times10^{-3}rad/s$), evaluated over a six second trajectory.
Hence, this tracking error indicates that the trajectory executed on the real robot closely matches ours optimised via latent-space control. 
The locomotion results are qualitatively more dynamic than traditional static walk trajectories, as the centre of mass of the robot is at rest for short periods of time and only during phases where all the robot's feet are in contact. %
This is an encouraging result, which means that a dynamic constraint predictor could produce faster trajectories.

\section{Conclusions}\label{section:Conclusions}

We propose an effective and novel approach to robot control based on high-level goal optimisation in a structured latent space, culminating in a demonstration of real-world quadruped locomotion. This is radically different from related works in that it directly exploits a statistical model of feasible robot configurations captured in a generative model to achieve smooth and realisable trajectories. These are able to stabilise a perturbed robot as well as optimise for stable and realisable locomotion — all while being computable an order of magnitude faster than equivalent analytical approaches. The optimisation is achieved via activation maximisation driven by performance predictors operating at a semantic level. This allows us to render complex, otherwise non-differentiable constraints directly usable in our approach. Our approach to latent-space control is compared with a feed-forward classifier network trained to predict stability directly from the input: the latter classifier also acting as a differentiable stability criterion. Compared to our approach, Input-AM is shown to both be significantly less successful at finding stable robot poses given an initially unstable pose and produces considerably fewer dynamically feasible trajectories than latent-AM. While we have shown the appeal and efficacy of performance predictors for applying complex non-differentiable constraints, we aim to investigate their robustness in future work. Finally, this approach and subsequent results shift the gaze from a traditional control viewpoint to a novel machine learning perspective leveraging deep generative models for complex, real-world robot control.

\section*{Acknowledgements}

This research was supported by an EPSRC Programme Grant [EP/M019918/1], by the UKRI/EPSRC RAIN Hub [EP/R026084/1] and the EPSRC grant ‘Robust Legged Locomotion’ [EP/S002383/1]. This work was conducted as part of ANYmal Research, a community to advance legged robotics.
The authors would like to thank Mathieu Geisert for his help with running the experiments on the real robot.
We would also like to thank  Oliwier Melon for a pipe-lining tool and Oliver Groth for his advice throughout the project.

\bibliographystyle{IEEEtran}
\bibliography{references}

\begin{thebibliography}{10}
\providecommand{\url}[1]{#1}
\csname url@rmstyle\endcsname
\providecommand{\newblock}{\relax}
\providecommand{\bibinfo}[2]{#2}
\providecommand\BIBentrySTDinterwordspacing{\spaceskip=0pt\relax}
\providecommand\BIBentryALTinterwordstretchfactor{4}
\providecommand\BIBentryALTinterwordspacing{\spaceskip=\fontdimen2\font plus
\BIBentryALTinterwordstretchfactor\fontdimen3\font minus
  \fontdimen4\font\relax}
\providecommand\BIBforeignlanguage[2]{{%
\expandafter\ifx\csname l@#1\endcsname\relax
\typeout{** WARNING: IEEEtran.bst: No hyphenation pattern has been}%
\typeout{** loaded for the language `#1'. Using the pattern for}%
\typeout{** the default language instead.}%
\else
\language=\csname l@#1\endcsname
\fi
#2}}

\bibitem{ANYmal}
M.~{Hutter}, C.~{Gehring}, D.~{Jud}, A.~{Lauber}, C.~D. {Bellicoso},
  V.~{Tsounis}, J.~{Hwangbo}, K.~{Bodie}, P.~{Fankhauser}, M.~{Bloesch},
  R.~{Diethelm}, S.~{Bachmann}, A.~{Melzer}, and M.~{Hoepflinger}, ``{ANYmal -
  A Highly Mobile and Dynamic Quadrupedal Robot},'' in \emph{IEEE/RSJ
  International Conference on Intelligent Robots and Systems (IROS)}, 2016.

\bibitem{robot_terrain_anymal}
P.~{Fankhauser}, M.~{Bjelonic}, C.~{Dario Bellicoso}, T.~{Miki}, and
  M.~{Hutter}, ``{Robust Rough-Terrain Locomotion with a Quadrupedal Robot},''
  in \emph{IEEE International Conference on Robotics and Automation (ICRA)},
  2018.

\bibitem{hyq}
C.~Semini, N.~Tsagarakis, E.~Guglielmino, M.~Focchi, F.~Cannella, and
  D.~Caldwell, ``{Design of HyQ - A Hydraulically and Electrically Actuated
  Quadruped Robot},'' \emph{Proceedings of the Institution of Mechanical
  Engineers. Part I: Journal of Systems and Control Engineering}, 2011.

\bibitem{cheatah_mini}
G.~{Bledt}, M.~J. {Powell}, B.~{Katz}, J.~{Di Carlo}, P.~M. {Wensing}, and
  S.~{Kim}, ``{MIT Cheetah 3: Design and Control of a Robust, Dynamic Quadruped
  Robot},'' in \emph{IEEE/RSJ International Conference on Intelligent Robots
  and Systems (IROS)}, 2018.

\bibitem{anymal_offshore}
C.~Gehring, P.~Fankhauser, L.~Isler, R.~Diethelm, S.~Bachmann, M.~Potz,
  L.~Gerstenberg, and M.~Hutter, ``\BIBforeignlanguage{en}{{ANYmal in the Field
  : Solving Industrial Inspection of an Offshore HVDC Platform with a
  Quadrupedal Robot}},'' 2019-08-28, 12th Conference on Field and Service
  Robotics (FSR 2019); Conference Location: Tokyo, Japan; Conference Date:
  August 29-31, 2019.

\bibitem{towr}
A.~W. {Winkler}, C.~D. {Bellicoso}, M.~{Hutter}, and J.~{Buchli}, ``{Gait and
  Trajectory Optimization for Legged Systems Through Phase-Based End-Effector
  Parameterization},'' \emph{IEEE Robotics and Automation Letters}, vol.~3,
  no.~3, pp. 1560--1567, July 2018.

\bibitem{DarioNlp}
C.~D. {Bellicoso}, F.~{Jenelten}, C.~{Gehring}, and M.~{Hutter}, ``{Dynamic
  Locomotion Through Online Nonlinear Motion Optimization for Quadrupedal
  Robots},'' \emph{IEEE Robotics and Automation Letters}, vol.~3, no.~3, pp.
  2261--2268, July 2018.

\bibitem{towr++}
O.~Melon, M.~Geisert, D.~Surovik, I.~Havoutis, and M.~Fallon, ``{Reliable
  Trajectories for Dynamic Quadrupeds using Analytical Costs and Learned
  Initializations},'' 2020.

\bibitem{fastTraj}
A.~W. {Winkler}, F.~{Farshidian}, D.~{Pardo}, M.~{Neunert}, and J.~{Buchli},
  ``{Fast Trajectory Optimization for Legged Robots Using Vertex-Based ZMP
  Constraints},'' \emph{IEEE Robotics and Automation Letters}, vol.~2, no.~4,
  pp. 2201--2208, Oct 2017.

\bibitem{static_stability}
T.~{Bretl} and S.~{Lall}, ``{Testing Static Equilibrium for Legged Robots},''
  \emph{IEEE Transactions on Robotics}, vol.~24, no.~4, pp. 794--807, Aug 2008.

\bibitem{vae}
D.~Kingma and M.~Welling, ``{Auto-Encoding Variational Bayes},'' in
  \emph{International Conference on Learning Representations (ICLR)}, 2014.

\bibitem{vae_1}
D.~J. Rezende, S.~Mohamed, and D.~Wierstra, ``{Stochastic Backpropagation and
  Approximate Inference in Deep Generative Models},'' in \emph{Proceedings of
  the International Conference on International Conference on Machine Learning
  (ICML)}, 2014.

\bibitem{imagine_that}
Y.~Wu, S.~Kasewa, O.~Groth, S.~Salter, L.~Sun, O.~{Parker Jones}, and
  I.~Posner, ``{Imagine That! Leveraging Emergent Affordances for Tool
  Synthesis in Reaching Tasks},'' \emph{arXiv preprint arXiv:1909.13561}, 2019.

\bibitem{activation_maximisation}
D.~Erhan, Y.~Bengio, A.~Courville, and P.~Vincent, ``{Visualizing Higher-Layer
  Features of a Deep Network},'' University of Montreal, Tech. Rep. 1341, June
  2009, also presented at the ICML 2009 Workshop on Learning Feature
  Hierarchies, Montr{\'{e}}al, Canada.

\bibitem{feasible_region}
R.~Orsolino, M.~Focchi, S.~Caron, G.~Raiola, V.~Barasuol, and C.~Semini,
  ``{Feasible Region: an Actuation-Aware Extension of the Support Region},''
  \emph{arXiv preprint arXiv: 1903.07999}, 2019.

\bibitem{centroidal_dynamics}
D.~Orin, A.~Goswami, and S.-H. Lee, ``{Centroidal Dynamics of a Humanoid
  Robot},'' \emph{Autonomous Robots}, vol.~35, 10 2013.

\bibitem{Jemin}
J.~Hwangbo, J.~Lee, A.~Dosovitskiy, D.~Bellicoso, V.~Tsounis, V.~Koltun, and
  M.~Hutter, ``{Learning Agile and Dynamic Motor Skills for Legged Robots},''
  \emph{CoRR}, vol. abs/1901.08652, 2019.

\bibitem{gangapurwala2020guided}
S.~Gangapurwala, A.~Mitchell, and I.~Havoutis, ``{Guided Constrained Policy
  Optimization for Dynamic Quadrupedal Robot Locomotion},'' 2020.

\bibitem{UPN}
A.~Srinivas, A.~Jabri, P.~Abbeel, S.~Levine, and C.~Finn, ``{Universal Planning
  Networks},'' in \emph{Proceedings of the International Conference on Machine
  Learning (ICML)}, 2018.

\bibitem{embed2control}
M.~Watter, J.~T. Springenberg, J.~Boedecker, and M.~A. Riedmiller, ``{Embed to
  Control: A Locally Linear Latent Dynamics Model for Control from Raw
  Images},'' in \emph{Advances in Neural Information Processing Systems
  (NeurIPS)}, 2015.

\bibitem{pca}
K.~P. F.R.S., ``{LIII. On Lines and Planes of Closest Fit to Systems of Points
  in Space}.''

\bibitem{SEA}
G.~A. {Pratt} and M.~M. {Williamson}, ``{Series Elastic Actuators},'' in
  \emph{IEEE/RSJ International Conference on Intelligent Robots and Systems
  (IROS)}, 1995.

\bibitem{elu}
D.-A. Clevert, T.~Unterthiner, and S.~Hochreiter, ``{Fast and Accurate Deep
  Network Learning by Exponential Linear Units (ELUs)},'' 01 2016.

\bibitem{wilson}
E.~B. Wilson, ``{Probable Inference, the Law of Succession, and Statistical
  Inference},'' \emph{Journal of the American Statistical Association},
  vol.~22, no. 158, pp. 209--212, 1927.

\bibitem{robust_TO}
M.~{Kalakrishnan}, J.~{Buchli}, P.~{Pastor}, M.~{Mistry}, and S.~{Schaal},
  ``{Fast, Robust Quadruped Locomotion over Challenging Terrain},'' in
  \emph{2010 IEEE International Conference on Robotics and Automation}, 2010,
  pp. 2665--2670.

\end{thebibliography}

\end{document}